\pdfoutput=1

\documentclass[11pt]{article}

\usepackage{times}
\usepackage{latexsym}
\usepackage{graphicx}
\usepackage{graphics}
\usepackage{wasysym}
\usepackage{lipsum}
\usepackage{supertabular}

\usepackage{acl}

\usepackage{times}
\usepackage{latexsym}

\usepackage[T1]{fontenc}

\usepackage[utf8]{inputenc}

\usepackage[T1]{fontenc}

\usepackage[utf8]{inputenc}

\title{Detecting Generated Scientific Papers using an Ensemble of Transformer Models}

\author{Anna Glazkova \\
  University of Tyumen  \\Tyumen, Russia \\
  \texttt{a.v.glazkova@utmn.ru} \\\And
  Maksim Glazkov \\
  Voctiv RnD d.o.o. Beograd \\ Belgrade, Serbia \\
  \texttt{my.eye.off@gmail.com} \\}

\begin{document}
\maketitle
\begin{abstract}
The paper describes neural models developed for the DAGPap22 shared task hosted at the Third Workshop on Scholarly Document Processing. This shared task targets the automatic detection of generated scientific papers. Our work focuses on comparing different transformer-based models as well as using additional datasets and techniques to deal with imbalanced classes. As a final submission, we utilized an ensemble of SciBERT, RoBERTa, and DeBERTa fine-tuned using random oversampling technique. Our model achieved 99.24\% in terms of F1-score. The official evaluation results have put our system at the third place.
\end{abstract}

\section{Introduction}

State-of-the-art natural language processing (NLP) tools generate high-quality texts that could hardly be distinguished from human-written texts. This represents a remarkable achievement in modern science, but raises challenges in terms of detecting machine-generated texts. Detection of automatically generated texts is crucial for many NLP tasks, in particular, for prevention of spreading fake scientific publications and citations \cite{else2021tortured}. Here we focus on the task of detecting automatically generated scientific excerpts as a part of the Third Workshop on Scholarly Document Processing shared tasks. The
source code that we used for fine-tuning our models as well as additional data generated by us are freely available\footnote{\url{https://github.com/oldaandozerskaya/DAGPap22}}.

The work is based on the participation of our team in the DAGPap22 shared task. The objective of the task is to detect automatically generated papers in terms of a binary classification task. This task is challenging due to the developing models for text generation and wide spreading of untruthful content on the internet. To date, language models for generating texts are widely used in the scientific domain, for example for producing long and short summaries \cite{gharebagh2020guir,cachola2020tldr,takeshita2022x}, citation texts \cite{xing-etal-2020-automatic,ge2021baco}, keyphrases \cite{glazkova2022applying,chowdhury2022applying}, peer reviews \cite{yuan2021can}. The scientific community has held several machine learning competitions to identify machine-generated texts in different domains \cite{uchendu2021turingbench,shamardina2022findings}. 

The paper is organized as follows. We provide the dataset and task description in Section \ref{sec:s1}. In Section \ref{sec:s2}, we describe our experiments during the development phase and report the official results. Section \ref{sec:s3} concludes this paper.

\section{Task Overview}\label{sec:s1}

\subsection{Task Definition}

The objective of the task is to identify whether a text is automatically generated. Therefore, the task represents a binary classification problem, the purpose of which is to split the given texts into two mutually exclusive classes. Formally, the problem is described as follows.
\begin{itemize}
    \item \textbf{Input.} Given a scientific excerpt.
    \item \textbf{Output.} One of two different labels, such as "human-written" or "machine-generated".
\end{itemize}

\subsection{Data}

The original training set contains 5350 excerpts from a scientific papers, among which 1686 are human-written and 3664 are machine-generated. The test set includes 21403 excerpts. The text corpus is based on the work by \citet{cabanac2021tortured}, as well as fragments collected by Elsevier publishing and editorial teams. The statistics is presented in Table \ref{table:stat}\footnote{The number of words and sentences was defined using NLTK \cite{bird2006nltk}}. Table \ref{table:a} contains some examples of automatically generated texts.

\begin{table}[h!]
\centering
\begin{tabular}{lll}
\hline
\multicolumn{1}{c}{\textbf{Characteristic}} & \multicolumn{1}{c}{\textbf{Train}} & \multicolumn{1}{c}{\textbf{Test}} \\ \hline
Avg number of words & 157.4 & 158.37 \\ 
Min number of words & 51 & 51 \\
Max number of words & 1895 & 1784 \\ 
Avg number of sentences & 5.8 & 5.75 \\ 
Min number of sentences & 1 & 1 \\ 
Max number of sentences & 63 & 68 \\ \hline
\end{tabular}
\caption{\label{font-table} Data statistics.}
\label{table:stat}
\end{table}

\begin{table}[h!]
\centering
\begin{tabular}{p{0.4cm}p{6.4cm}}
\hline
\multicolumn{1}{c}{\textbf{ID}} & \multicolumn{1}{c}{\textbf{Excerpt}} \\ \hline
23& Electronic nose or machine olfaction are systems used for detection and identification of odorous compounds and gas mixtures Electronic nose or machine olfaction are systems used for detection and identification of odorous compounds and gas mixtures. Olfactors, e.g. motorbikes, are used for odor detection. These devices do not detect volatile agents or gas mixtures, and cannot be used for quantitative odor determination.\\\hline
55 & For the low price of coal and ineffective environmental management in mining area, China is in the dilemma of the increasing coal demand and the serious environmental issues in mining area For the low price of coal and ineffective environmental management in mining area, China is in the dilemma of the increasing coal demand and the serious environmental issues in mining area.\\\hline
242 & The motivation behind this paper is to answer analysis of the past portrayals of Sandler and Smith of the numeraire in an intertemporal investigation of Pareto effectiveness conditions. This reevaluation recommends that the job of the numeraire is demonstrated to be less obvious than Cabe infers. In addition, the examination shows that the prior ends are not critically subject to the numeraire presumption.\\\hline
\end{tabular}
\caption{\label{font-table} Examples of generated texts from the official training set.}
\label{table:a}
\end{table}

\section{Our Work}\label{sec:s2}

\subsection{Models}

\begin{table}[h]
\centering
\begin{tabular}{ll}
\multicolumn{1}{c}{\textbf{Model}} & \multicolumn{1}{c}{\textbf{Value}} \\\hline
\multicolumn{2}{c}{\textbf{Vocabulary (K)}} \\\hline
SciBERT & 30 \\
RoBERTa & 50 \\
DeBERTa & 50 \\\hline

\multicolumn{2}{c}{\textbf{Backpone Parameteres (M)}} \\\hline
SciBERT & 110 \\
RoBERTa & 355 \\
DeBERTa & 350 \\\hline

\multicolumn{2}{c}{\textbf{Hidden Size}} \\\hline
SciBERT & 768 \\
RoBERTa & 1024 \\
DeBERTa & 1024 \\\hline

\multicolumn{2}{c}{\textbf{Layers}} \\\hline
SciBERT & 12 \\
RoBERTa & 16 \\
DeBERTa & 24 \\\hline

\end{tabular}
\caption{\label{font-table} Hyperparameteres of the considered BERT-based models (SciBERT$_{base-cased}$, RoBERTa$_{large}$, and DeBERTa$_{large}$).}
\label{table:hyp}
\end{table}

\begin{table*}[h!]
\centering
\begin{tabular}{llll}
\hline
\multicolumn{1}{c}{\textbf{Model}} & \multicolumn{1}{c}{\textbf{P}} & \multicolumn{1}{c}{\textbf{R}} & \textbf{F1} \\ \hline
SciBERT$_{128}$ &  96.19&  94.69&  95.38\\ 
SciBERT$_{256}$ &  97.58&  96.49&  96.99\\ 
SciBERT$_{512}$ &  97.84& 97.16 &  97.49\\ 
RoBERTa &  96.54&  94.89&  95.65\\ 
DeBERTa & 97.35 & 97 & 97.17 \\ \hline
SciBERT$_{512}$ + oversampling &\textbf{98.2} & 97.92 &  \textbf{98.06}\\ 
SciBERT$_{512}$ + undersampling &  97.07&  95.42&  96.15\\ 
SciBERT$_{512}$ + class weighting &  98.05&  97.81&  97.93\\ 

RoBERTa + oversampling &  96.92&  96.5&  96.7\\ 
RoBERTa + undersampling &  95.55&  92.83&  93.89 \\ 
RoBERTa + class weighting & 96.62 & 96.49 & 96.56\\

DeBERTa + oversampling & 97.51 & 96.61 & 97.04 \\ 
DeBERTa + undersampling &  95.62&  93.04&  94.13\\

\hline
SciBERT$_{512}$ + KP20K (BT) + oversampling&  97.65&  \textbf{98.18}&  97.91\\ 
SciBERT$_{512}$ + KP20K (GPT-2) + oversampling&  97.16&  97.03&  97.07\\ 
SciBERT$_{512}$ + original (BT) + oversampling&  97.44&  97.75&  97.59\\ 
SciBERT$_{512}$ + original (GPT-2) + oversampling&  97.56&  98.15&  97.84\\

RoBERTa + KP20K (BT) + oversampling&  96.86&  96.48& 96.66 \\ 
RoBERTa + KP20K (GPT-2) + oversampling&  96.49&  95.2& 95.8 \\ 
RoBERTa + original (BT) + oversampling&  96.56& 95.99  & 96.26 \\ 
RoBERTa + original (GPT-2) + oversampling&  96.12&96.12  &96.12  \\ 

DeBERTa + KP20K (BT) + oversampling&  96.76&  97.03& 96.89 \\ 
DeBERTa + KP20K (GPT-2) + oversampling & 94.16 & 95.86 & 94.95 \\  
DeBERTa + original (BT) + oversampling&  96.51&  96.7& 96.59  \\ 
DeBERTa + original (GPT-2) + oversampling& 96.58 &96.94  &96.76  \\  \hline
\end{tabular}
\caption{\label{font-table} Results (\%, development phase).}
\label{table:res}
\end{table*}

In this work, we used neural models based on Bidirectional Encoder Representations from Transformers (BERT) \cite{devlin-etal-2019-bert} because they showed high results in the scientific domain \cite{glazkova2021identifying,pan2021bert,zhu2021bert}. We experimented with the following models, the overview of which is presented in Table \ref{table:hyp}:

\begin{itemize}
    \item SciBERT$_{base-cased}$ \cite{beltagy-etal-2019-scibert}, a BERT-based model that is pretrained on the texts of papers taken from Semantic Scholar.
    \item RoBERTa$_{large}$ \cite{liu2019roberta}, a modification of BERT that is pretrained using dynamic masking.
    \item DeBERTa$_{large}$ \cite{he2020deberta}, a model that is pretrained using disentangled attention and enhanced mask decoder.
\end{itemize}

To evaluate our models during the development phase, we performed 3-fold cross-validation on the training set. The results were evaluated in terms of macro-averaged F1-score (F1), precision (P), and recall (R).

\subsection{Experiments}

We adopted pretrained models from HuggingFace \cite{wolf2020transformers} and fine-tuned them using SimpleTransformers\footnote{\url{https://simpletransformers.ai}}. We fine-tuned each pre-trained language model for three epochs with the learning rate of 2e-5 using the AdamW optimizer \cite{loshchilov2017decoupled}. We set batch size to 16 and used the sliding window technique to prevent truncating longer sequences. We utilized the maximum sequence length equal to 128, 256, and 512 for SciBERT (SciBERT$_{128}$, SciBERT$_{256}$, and SciBERT$_{512}$ respectively) and 128 for RoBERTa and DeBERTa due to the limited computing resources. Similar to our previous work \cite{glazkova2021g2tmn}, we used raw texts as an input.

\begin{figure*}
    \centering
    \includegraphics[width=1\textwidth]{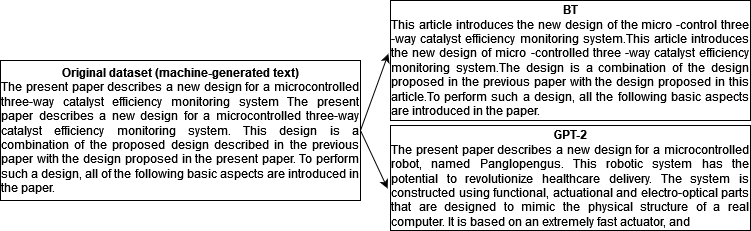}
    \caption{Example of generating new data using BT and GPT-2.}
    \label{fig:example}
\end{figure*}

Since the corpus provided by the organizers is imbalanced, we explored several techniques to handle imbalanced data. Namely, we used a) random oversampling, b) random undersampling, c) class weighting, d) generating new data. Random oversampling and undersampling are implemented using the Imbalanced-learn library\footnote{\url{https://imbalanced-learn.org}}. To generate new data, we experimented with the original corpus and the fragment of the KP20K dataset \cite{meng2017deep}. KP20K is a large-scale scholarly papers dataset for keyphrase extraction containing 568K papers with their abstracts. To produce new machine-generated data, we utilized two techniques for text generation: a) Back Translation (BT)\footnote{\url{https://github.com/hhhwwwuuu/BackTranslation}} through Googletrans\footnote{\url{https://py-googletrans.readthedocs.io/en/latest}}, and b) zero-shot generation by prompting GPT-2 \cite{radford2019language} and specifying the maximum number of generated tokens equal to the number of tokens in the source text (see Figure \ref{fig:example} for example).

The results are presented in Table \ref{table:res}. In our experiments, the model fine-tuned on longer input sequences (SciBERT$_{512}$) performed better than other baselines despite the use of the sliding window technique. Due the processing of class imbalance, we found that oversampling and class weighting increase the performance of the models while undersampling produces lower results. Further, we experimented with using additional data. First, we made an attempt to add scientific abstracts from KP20K utilizing texts of 1000 random abstracts and 1000 texts generated by BT or GPT-2 and then perform oversampling. Second, we tried to produce new examples of machine-generated excerpts from the dataset provided by the organizers of the competition. We generated 1000 examples using BT and GPT-2, added them to the training set, and finally performed oversampling. The use of additional data showed no increase compared to the models fine-tuned with oversampled texts.

\subsection{Results}

\begin{figure}
    \centering
    \includegraphics[width=0.5\textwidth]{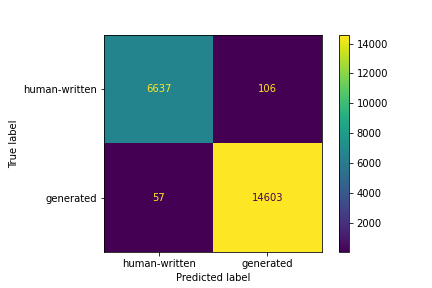}
    \caption{Confusion matrix for our model.}
    \label{fig:cm}
\end{figure}

During the evaluation phase, we experimented with the hard and soft voting ensembles of transformer-based models. The results were evaluated on the official test set. Our best submission is an ensemble of SciBERT, RoBERTa, and DeBERTa fine-tuned using random oversampling technique. The confusion matrix for this solution is presented in Figure \ref{fig:cm}. The ensembling of predictions was performed at two levels:

\begin{enumerate}
    \item Model level, i. e. soft voting calculated for three models of the same type fine-tuned with different random seeds.
    \item Ensemble level, i. e. hard voting for the labels produced by the models of different type.
\end{enumerate}

Table \ref{table:res1} shows the comparison of our best solution to the official scores from the private leaderboard of the competition\footnote{\url{https://www.kaggle.com/competitions/detecting-generated-scientific-papers}}. In this competition, only five models outperformed the baseline provided by the organizers. Our model achieved 99.24\% of F1-score and ranked the third place of the leaderboard for this task.

\begin{table}[h!]
\centering
\begin{tabular}{ll}
\hline
\multicolumn{1}{c}{\textbf{Run name}} & \multicolumn{1}{c}{\textbf{F1}} \\ \hline
Our solution & 99.24 \\
Stronger benchmark & 98.32\\
Tf-Idf \& logreg benchmark & 82.04\\\hline
Average scores &  92.96 \\\hline
\end{tabular}
\caption{\label{font-table} Official results (\%, private leaderboard).}
\label{table:res1}
\end{table}

\section{Conclusion}\label{sec:s3}

In this work, we have explored the application of BERT-based models to the task of detecting machine-generated scientific texts. We have evaluated several techniques for handling imbalanced data and compared three models in a variety of settings. Our results on the test data showed that the ensemble of different transformer-based models outperforms other our submissions and strong baselines. Moreover, our final model ranked third in this task.

A further study could explore the state-of-the art-in detecting automatically generated papers for other languages and multilingual corpora. Another future direction is to continue our experiments with generating new data to improve the classification performance.

\bibliography{anthology,acl_latex}
\bibliographystyle{acl_latex}

\end{document}